\documentclass[11pt,a4paper]{article}
\usepackage[hyperref]{emnlp-ijcnlp-2019}
\usepackage{times}
\usepackage{latexsym}

\usepackage{url}
\usepackage{multirow}
\usepackage{algorithm}
\usepackage{algorithmic}
\usepackage{graphicx}
\usepackage{amsmath}
\usepackage{amsfonts}

\aclfinalcopy 

\title{Representation Learning with Ordered Relation Paths \\ for Knowledge Graph Completion}

\author{Yao Zhu$^1$, Hongzhi Liu$^2$\thanks{\, Corresponding author} , Zhonghai Wu$^{3,4}$, Yang Song$^5$ \and Tao Zhang$^5$ \\
  $^1$ Center for Data Science, Peking University, Beijing, China \\
  $^2$ School of Software and Microelectronics, Peking University, Beijing, China \\
  $^3$ National Engineering Center of Software Engineering, Peking University, Beijing, China \\
  $^4$ Key Lab of High Confidence Software Technologies (MOE), Peking University, Beijing, China \\
  $^5$ BossZhipin NLP Center @Kanzhun.com, Beijing, China \\
  {\tt \{yao.zhu,liuhz,wuzh\}@pku.edu.cn, \{songyang,kylen.zhang\}@kanzhun.com} \\}

\date{}

\begin{document}
\maketitle

\begin{abstract}
Incompleteness is a common problem for existing knowledge graphs
(KGs), and the completion of KG which aims to predict links between
entities is challenging. Most existing KG completion methods only
consider the direct relation between nodes and ignore the relation
paths which contain useful information for link prediction.
Recently, a few methods take relation paths into consideration but
pay less attention to the order of relations in paths which is
important for reasoning. In addition, these path-based models always
ignore nonlinear contributions of path features for link prediction. To solve these
problems, we propose a novel KG completion method named OPTransE.
Instead of embedding both entities of a relation into the same
latent space as in previous methods, we project the head entity and
the tail entity of each relation into different spaces to guarantee
the order of relations in the path. Meanwhile, we adopt a pooling strategy to
extract nonlinear and complex features of different paths to
further improve the performance of link prediction. Experimental
results on two benchmark datasets show that the proposed model
OPTransE performs better than state-of-the-art methods.
\end{abstract}

%
\maketitle

\section{Introduction}

Knowledge graphs (KGs) are built to store structured facts which are
encoded as triples, e.g., (\textbf{Beijing}, \textsl{CapitalOf},
\textbf{China}) \cite{lehmann2015dbpedia}. Each triple $(h,r,t)$
consists of two entities ${h}$, ${t}$ and a relation ${r}$,
indicating there is a relation ${r}$ between ${h}$ and ${t}$.
Large-scale KGs such as YAGO \cite{suchanek2007yago}, Freebase
\cite{bollacker2008freebase} and WordNet \cite{miller1995wordnet}
contain billions of triples and have been widely applied in various
fields \cite{riedel2013relation,dong2015question}.
However, a common problem with these KGs is that they are far from
complete, which has limited the development of KG's applications.
Thus, KG completion with the goal of filling in missing parts of the KG
has become an urgent issue. Specifically, KG completion aims to
predict whether a relationship between two entities is likely to be
true, which is defined as the link prediction in KGs.

Most existing KG completion methods are based on representation
learning, which embed both entities and relations into continuous
low-dimension spaces. TransE \cite{bordes2013translating} is one of
the most classical KG completion models, which embeds entities and
relations into the same latent space. To better deal with complex
relations like 1-to-N, N-to-1 and N-to-N, TransH
\cite{wang2014knowledge} and TransR \cite{lin2015learning} employ
relation-specific hyperplanes and relation-specific spaces respectively to
separate triples according their corresponding relation. Unfortunately, these
models ignore the relation paths between entities which are helpful
for reasoning. For example, if we know A is B's brother, and B is
C's parent, then we can infer that A is C's uncle.

\begin{figure}[]
    \centering
    \includegraphics[width=1.0\linewidth]{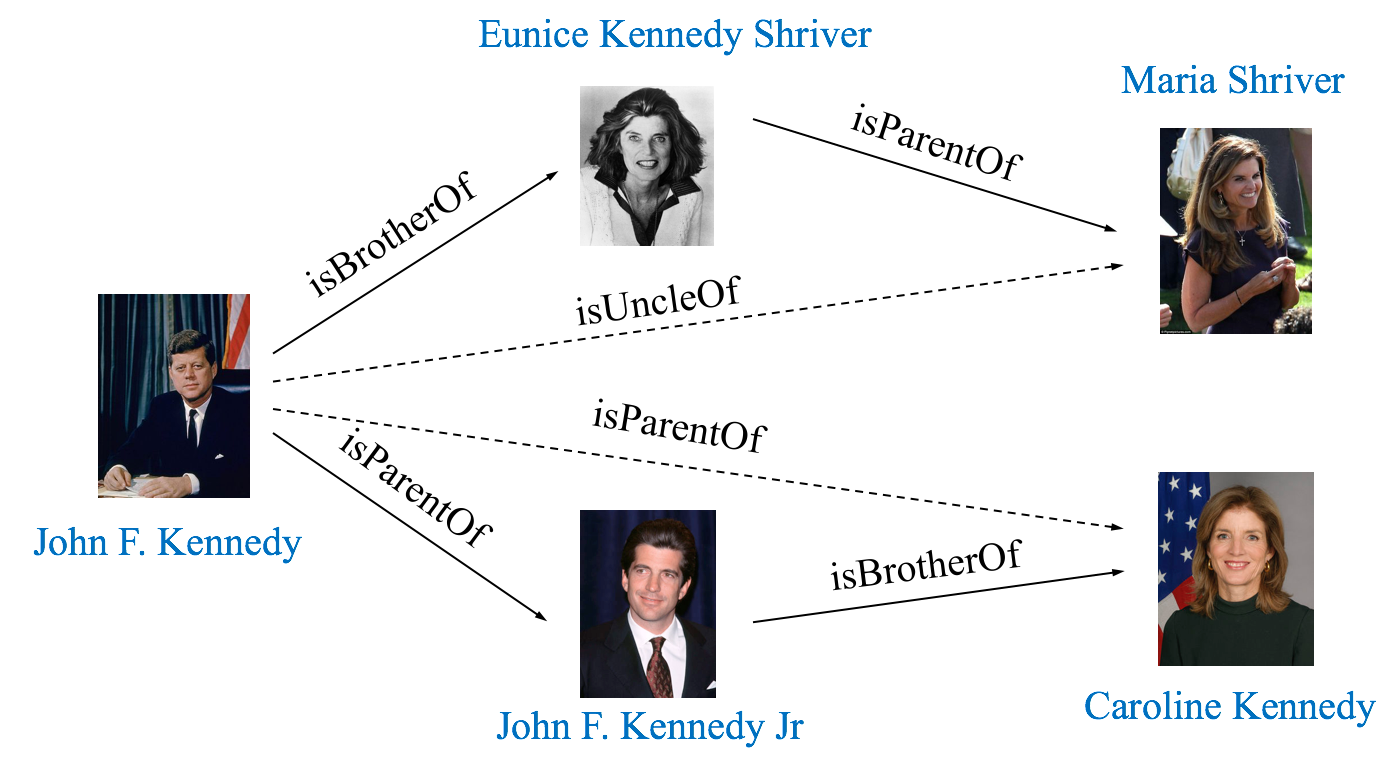}
    \caption{Example of the meaning change when the order of relations is altered.}
    \label{example}
\end{figure}

Recently, a few researchers take relation paths in KGs as additional
information for representation learning and attempt to project paths
into latent spaces, which get better performance compared with
conventional methods. PTransE-ADD \cite{lin2015modeling} considers
relation paths as translations between entities and represents each
path as the vector sum of all the relations in the path. Moreover,
RPE \cite{lin2018relation} extends the TransR model by incorporating
the path-specific projection. However, these methods pay less
attention to the order of relations in paths which is important for
link prediction. Figure \ref{example} shows an example of the
meaning change when the order of relations is altered. In
addition, these path-based models assume information from different
paths between an entity pair only contributes to the relation inference linearly
and ignore other complex interactions between them.

To address these issues, we propose a novel KG completion model
named OPTransE. In the model, we project the head entity and the
tail entity of each relation into different spaces and introduce sequence matrices to keep
the order of relations in the path. Moreover, a pooling strategy is adopted to
extract nonlinear features of different paths for relation
inferences. Experimental results on two benchmark datasets WN18 and
FB15K show that OPTransE significantly outperforms state-of-the-art
methods.

The remainder of this paper is organized as follows. Section 2
discusses related work. Section 3 presents the proposed model and
algorithm in detail. Empirical evaluation of the proposed algorithm
and comparison with other state-of-the-art algorithms are presented
in Section 4. Finally, Section 5 summarises the whole paper and
points out some future work.

\section{Related Work}

\subsection{Translation-based Models}

In recent years, there has been a great deal of work on
representation learning for KG completion, and most studies
concentrate on translation-based models. This kind of models
propose to embed both entities and relations into a continuous
low-dimensional vector space according to some distance-based
scoring functions.

TransE \cite{bordes2013translating} is one of the most fundamental
and representative translation-based models. For the entities and
relations in KGs, TransE encodes them as vectors in the same space.
For each fact $(h, r, t)$, TransE believes that \textbf{h} +
\textbf{r} $\approx$ \textbf{t} when $(h, r, t)$ holds. Thus, the
scoring function is defined as
\begin{equation}
f_{r}\left(h,t\right) = \parallel \mathbf{h} + \mathbf{r} -
\mathbf{t} \parallel_{1/2}.
\end{equation}
where \textbf{h}, \textbf{r} and \textbf{t} represent the vectors of
head entity $h$, relation $r$ and tail entity $t$, respectively. If
the fact $(h, r, t)$ is true, its score $f_{r}\left(h,t\right)$
tends to be close to zero.

TransE is a simple and efficient method for KG completion. However,
its simple structure has flaws in dealing with complicated relations
like 1-to-N, N-to-1 and N-to-N.
In order to address this problem, TransH
\cite{wang2014knowledge} introduces relation-specific hyperplanes
and projects entities as vectors onto the given hyperplanes. Similar
to TransH, TransR \cite{lin2015learning} also aims to cope with
complicated relations. Instead of employing the hyperplane like TransH, TransR proposes a
matrix $\mathbf{W}_{\mathbf{r}} \mathbf{\in}
\mathbb{R}^{m{\times}n}$ to project entity vectors into a
relation-specific space. Moreover, STransE \cite{nguyen2016stranse}
extends TransR by introducing two projection matrices for the head
entity and the tail entity, respectively. Therefore, the head and
tail entities in a triple will be projected differently into the
corresponding relation space.

\begin{figure*}[]
    \centering
    \includegraphics[width=1.0\linewidth]{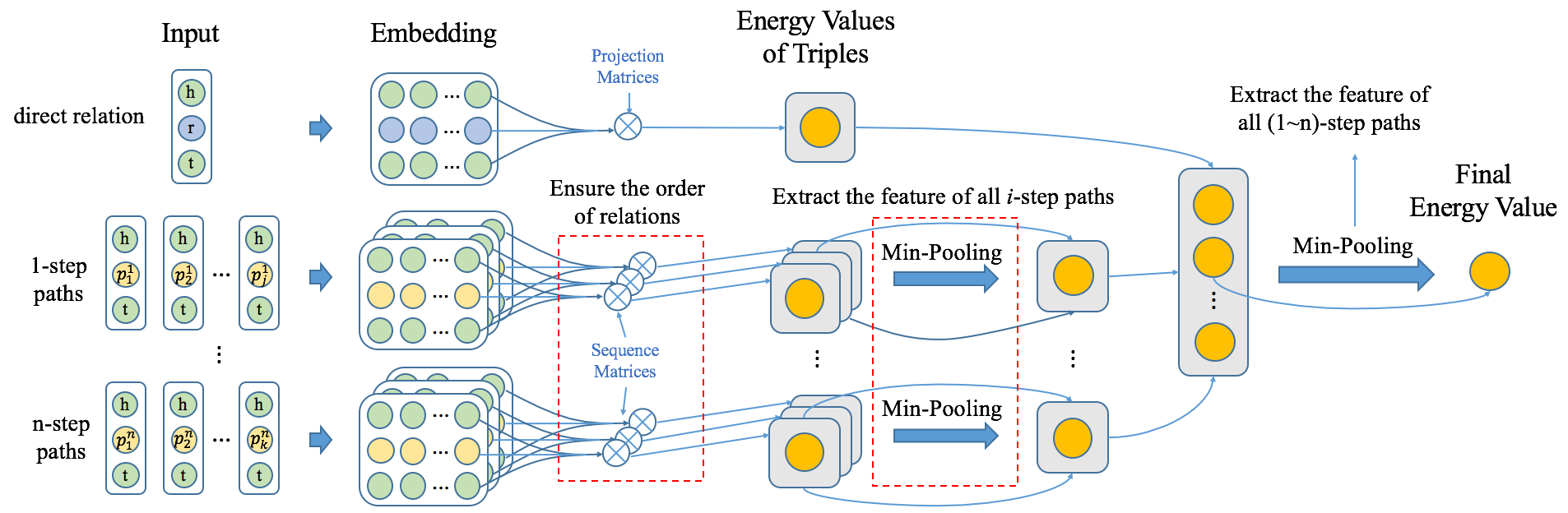}
    \caption{Architecture of OPTransE.}
    \label{Architecture}
\end{figure*}

\subsection{Incorporating Relation Paths}

The models introduced so far only exploit facts observed in KGs to
conduct representation learning. In fact, there is a large amount of
useful information in relation paths that can be incorporated into
translation-based models to improve the performance of link
prediction.

Lin et al. \shortcite{lin2015modeling} proposes a path-based
translation model named PTransE for KG completion. It regards
relation paths as translations between entities for representation
learning and utilizes a path-constraint resource allocation
algorithm to evaluate the reliability of relation paths. RTransE
\cite{garcia2015composing} and TransE-COMP \cite{guu2015traversing}
take the sum of the vectors of all relations in a path as the
representation for a relation path. For the Bilinear-COMP model
\cite{guu2015traversing}, and the PRUNED-PATHS model
\cite{toutanova2016compositional}, they represent each relation as a
diagonal matrix, and evaluate the relation path by matrix
multiplication. Most recently, PaSKoGE model \cite{jia2018path} is
proposed for KG embedding by minimizing a path-specific margin-based
loss function. Moreover, RPE \cite{lin2018relation}, inspired by
PTransE, extends the TransR model by incorporating the path-specific
projection for paths between entity pairs.

These methods try to incorporate information of relation paths to
get better performance. However, they pay less attention to the
order of relations in a path when learning representations of the
path. In fact, changes in the relation order of paths will alter the
meanings of paths to a great extent (as shown in Figure
\ref{example}). Moreover, the methods stated above assume
information from different paths between an entity pair only contributes to the relation inference linearly. Unfortunately, they ignore the
complex nonlinear features of different paths. In order to solve
these problems, we propose OPTransE, a novel KG completion model,
which learns representations of ordered relation paths and designs a
pooling method to better extract nonlinear features from various
relation paths.

\section{Our Model}

To infer the missing parts of KGs, we propose a KG completion model
called OPTransE, whose architecture is shown in Figure
\ref{Architecture}. We first embed the entities and relations of KG
into latent spaces with the consideration of the order of relations
in paths. Then, we try to infer the missing relations using these
latent representations. Different from previous methods which embed
the head and tail of a relation into the same latent space, we
project them into different spaces. Therefore, we can distinguish
the order of relations in the path. To extract the complex and nonlinear
path information for relation reasoning, we design a two layer pooling
strategy to fuse the information from different paths.

In this section, we will first introduce the embedding
representations of ordered relation paths. After that, we utilize a
two layer pooling strategy to construct the total energy function of
triples and then the objective function is presented. Finally, we
will describe the detail of model implementation and analyze the complexity of the model.

\subsection{Ordered Relation Paths Representation}

For each triple $(h,r,t)$ in KG, we employ vectors to represent the
entity pair and the relation. Specifically,
\(\mathbf{h}\mathbf{\in}\mathbb{R}^{{d}}\) denotes the head entity
$h$, \(\mathbf{t}\mathbf{\in}\mathbb{R}^{{d}}\) denotes the tail
entity $t$ and \(\mathbf{r}\mathbf{\in}\mathbb{R}^{{d}}\) indicates
the relation $r$.

We assume the paths connecting two entities contain indicative
information for the direct relation between these two entities. To
measure these kinds of indicative effects while guarantee the order
of relations in a path, we define an energy function in Equation
(\ref{multi-step}). Let \(p^{s = n} \) denote one of the $n$-step
path from ${h}$ to ${t}$, i.e., ${h}\stackrel{
{r\textsubscript{1}}}{\longrightarrow}\cdots\stackrel{
{r\textsubscript{\emph{n}}}}{\longrightarrow}{t}$. If the relation
path is reasonable from ${h}$ to ${t}$, it will obtain a lower
energy value.
\begin{equation}
E \left( h,p^{s = n},t \right) =
\parallel \mathbf{h_{p}} + \sum\limits_{{i = 1}}^{{n}}\mathbf{S}^{i}_{\mathbf{p}}\mathbf{r}_{i} - \mathbf{t_{p}}
\parallel_{1/2},
\label{multi-step}
\end{equation}
where
\begin{equation}
\mathbf{h_{p}} = f({p},\mathbf{h}), \quad \mathbf{t_{p}} =
g({p},\mathbf{t}), \label{fg}
\end{equation}
\begin{equation}
\mathbf{S}^{i}_{\mathbf{p}} = \mathbf{W}({p, i}). \label{w}
\end{equation}

$\mathbf{h_{p}}$ and $\mathbf{t_{p}}$ denote the representations of
the head entity $h$ and the tail entity $t$ in the ordered relation
path $p$, respectively.
$\mathbf{S}^{i}_{\mathbf{p}}\mathbf{\in}\mathbb{R}^{{d \times}{d}}$ denotes
the sequence matrix with respect to the $i$-th relation in the given
path $p$.

\begin{figure*}[]
    \centering
    \includegraphics[width=0.95\linewidth]{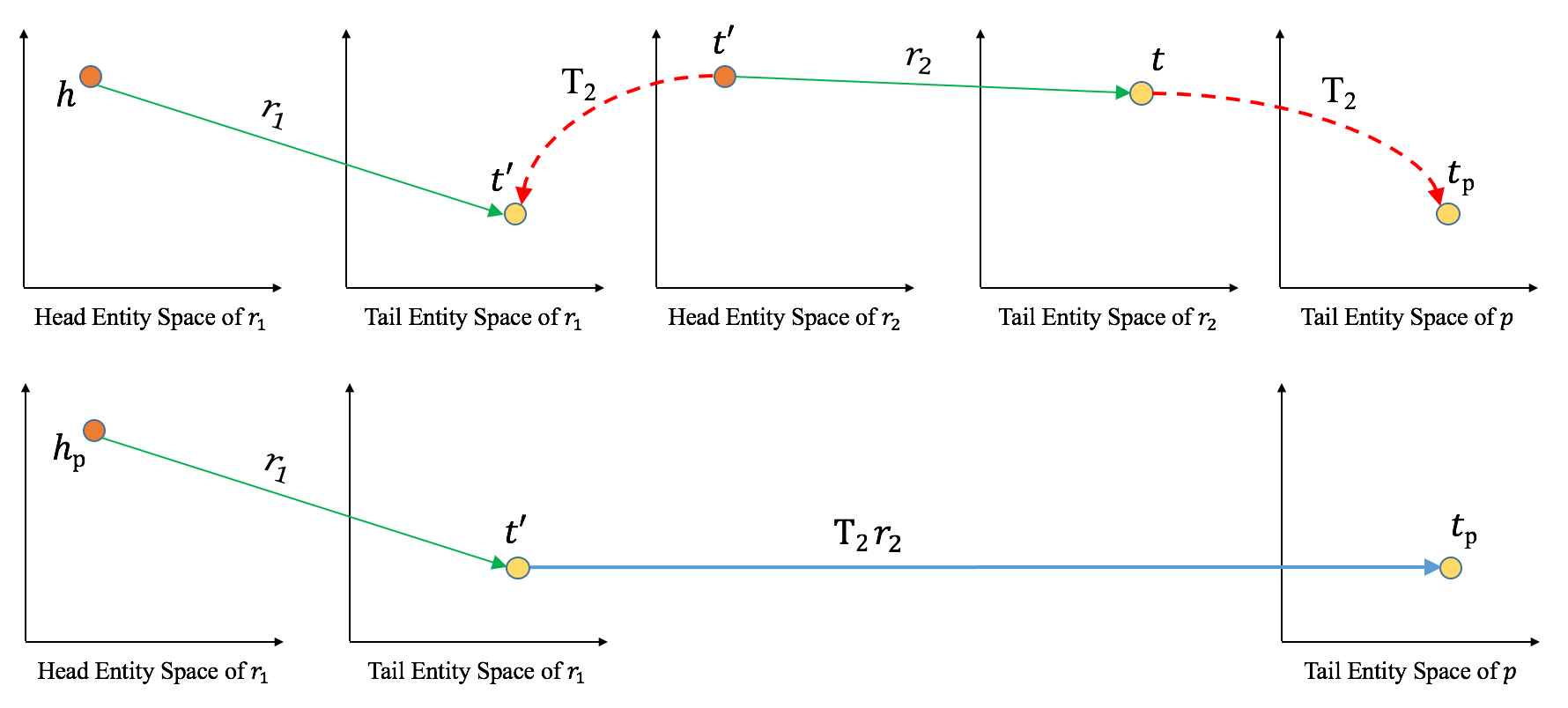}
    \caption{The representation of path $p$ (${h}\stackrel{
{r\textsubscript{1}}}{\longrightarrow}{t'}\stackrel{
{r\textsubscript{\emph{2}}}}{\longrightarrow}{t})$. The top part of the figure depicts the process of space transition of $t'$ and $t$, while the bottom part illustrates the generated continuous path from $h_{p}$ to $t_{p}$  after the transition.} 
\label{path}
\end{figure*}

Note that a triple $(h,r,t)$ in the KG can be seen as a one-step
path between $h$ and $t$. Thus, the value of $E(h,r,t)$ is able to
be obtained by substituting direct relation $r$ as $p^{s=1}$ into
Equation (\ref{multi-step}).

From Equation (\ref{multi-step}) we could observe that the sequence
matrix $\mathbf{S}^{i}_{\mathbf{p}}$ before each relation
$\textbf{r}_{i}$ is different. If the order of several relations in
a path is altered, the value of energy function will also change at
the same time. Therefore, paths with the same relation set but
different relation order will infer out distinct direct relations in
our model. The specific representation of the ordered relation path
will be demonstrated in the following contents.

To keep the order information of relations in paths, we project the
head and tail entities of a relation into different spaces by
introducing two matrices for each relation. Let
\(\mathbf{W}_{\mathbf{r,1}}\mathbf{\in}\mathbb{R}^{{d \times}{d}}\)
and \(\mathbf{W}_{\mathbf{r,2}}\mathbf{\in}\mathbb{R}^{{d
\times}{d}}\) denote the projection matrices of the head entity and
the tail entity for relation ${r}$, respectively. With these two
matrices, we will project the head and tail entities into distinct
spaces with respect to the same relation. Suppose there is a
path ${r\textsubscript{1}}, {r\textsubscript{2}}, \ldots{},
{r\textsubscript{n}}$ from ${h}$ to ${t}$, ideally, we define the
following equations

\begin{equation}
\left\{ \begin{matrix}
\mathbf{W}_{\mathbf{r}_{\mathbf{1}}\mathbf{,1}}\mathbf{h} + \mathbf{r}_{\mathbf{1}} = \mathbf{W}_{\mathbf{r}_{\mathbf{1}}\mathbf{,2}}\mathbf{t}^{{(1)}}\mathbf{\text{\ \ \ \ }} \\
\mathbf{W}_{\mathbf{r}_{\mathbf{2}}\mathbf{,1}}\mathbf{t}^{{(1)}} + \mathbf{r}_{\mathbf{2}} = \mathbf{W}_{\mathbf{r}_{\mathbf{2}}\mathbf{,2}}\mathbf{t}^{{(2)}} \\
 \vdots \\
\mathbf{W}_{\mathbf{r}_{\mathbf{n}}\mathbf{,1}}\mathbf{t}^{{(n -
1)}} + \mathbf{r}_{\mathbf{n}} =
\mathbf{W}_{\mathbf{r}_{\mathbf{n}}\mathbf{,2}}\mathbf{t}
\label{matrix_eq}
\end{matrix} \right.\
,
\end{equation}
\newline
where \(\mathbf{t}^{{(i)}}\) indicates the $i$-th passing node on
the path.

For the entity pair with a relation path, we get their
representations after eliminating the passing nodes from Equation
(\ref{matrix_eq}). Thus, the concrete forms of the variables in
Equation (\ref{multi-step}) are shown as follows,
\begin{equation}
\mathbf{h_{p}} =
\mathbf{W}_{\mathbf{r}_{\mathbf{1}}\mathbf{,1}}\mathbf{h}, \quad
\mathbf{t_{p}} =  \mathbf{W}_{\mathbf{p^{s=n}}}\mathbf{t},
\label{hptp}
\end{equation}

\begin{equation}
\mathbf{S}^{i}_{\mathbf{p}} = \prod\limits_{{k =
1}}^{{i}}\mathbf{T}_{{k}},
\end{equation}

where
\begin{equation}
\mathbf{W}_{\mathbf{p^{s=n}}} = \mathbf{S}^{n}_{\mathbf{p}}
\mathbf{W}_{\mathbf{r}_{\mathbf{n}}\mathbf{,2}}, \label{wp}
\end{equation}

\begin{equation}
\mathbf{T}_{k} = \left\{
\begin{aligned}
& \qquad \quad \mathbf{\emph{I}}  & k = 1\\
& \mathbf{M}\left(r_{k}, r_{k-1}\right) \quad & k > 1
\end{aligned}
\right. . \label{Tk}
\end{equation}

$\mathbf{W}_{\mathbf{p^{s=n}}}\mathbf{\in}\mathbb{R}^{{d \times}{d}}$ indicates the projection matrix for
path $p^{s=n}$, which aims to project the tail entity in a path to
the space of $p^{s=n}$. Moreover, \emph{I} in Equation (\ref{Tk})
denotes the identity matrix and $\mathbf{M}\left(r_{k},
r_{k-1}\right)\mathbf{\in}\mathbb{R}^{{d \times}{d}}$ means the space transition matrix from the head
entity space of $r_{k}$ to the tail entity space of $r_{k-1}$, i.e., $\mathbf{M}\left(r_{k}, r_{k-1}\right)  \mathbf{W}_{\mathbf{r}_{\mathbf{k}}\mathbf{,1}} = \mathbf{W}_{\mathbf{r}_{\mathbf{k-1}}\mathbf{,2}}$ .

Figure \ref{path} illustrates the representation of the relation
path in our model. Suppose there is a 2-step path from ${h}$ to
${t}$ passing ${t'}$, i.e., ${h}\stackrel{
{r\textsubscript{1}}}{\longrightarrow}{t'}\stackrel{
{r\textsubscript{2}}}{\longrightarrow}{t}$. It is obvious that
${t'}$ acts as the tail entity of relation ${r\textsubscript{1}}$
and as the head entity of relation ${r\textsubscript{2}}$ at the
same time, which is shown on the top part of Figure \ref{path}. To
connect relations in different spaces, we try to unify the passing
node in the path into the same space. As defined in Equation
(\ref{Tk}), $T_{2}$ is utilized to transfer the passing node $t'$
from the head entity space of $r_{2}$ to the tail entity space of
$r_{1}$. Moreover, $T_{2}$ is also assigned to the relation $r_{2}$
and the tail entity $t$. Note that the tail entity $t$ will be
projected into the space of path $p$ which is defined in Equation
(\ref{hptp}). Finally, the path from $h_{p}$ to $t_{p}$ will pass
through $r_{1}$ and $T_{2}r_{2}$ as shown on the bottom part of
Figure \ref{path}.

\subsection{Pooling Strategy}
We design a two layer pooling strategy to fuse the information from
different paths. First, we utilize a minimum pooling method to
extract feature information from paths with $i$ steps and define an
energy function as follows,
\begin{equation}
\resizebox{1\linewidth}{!}{$ E\left( h,P^{s = i}_{r},t \right) =
\textrm{Min}\lbrack\ E\left( h,p^{s = i},t \right)|\ p^{s = i} \in
P^{s = i}_{r}\rbrack,$}
\end{equation}
where \(P^{s = i}_{r}\) indicates the set of all \emph{i}-step
paths which are relevant to the relation ${r}$  from the head entity ${h}$ to the tail entity ${t}$. To obtain \(P^{s = i}_{r}\), we introduce a conditional probability
\(\Pr(r|p^{s = i})\) to represent the reliability of a path \(p^{s =
i}\) associated with the given relation ${r}$,
\begin{equation}
\begin{split}
\Pr\left( r \middle| p^{s = i} \right) & = \Pr(r,p^{s = i})/\ \Pr(p^{s
= i}) \\
& = \frac{N(r, p^{s=i}) / N(p)}{N(p^{s=i}) / N(p)} \\
& = \frac{N(r, p^{s=i})}{N(p^{s=i})},
\end{split}
\end{equation}
where $\Pr(r,p^{s = i})$ denotes the joint probability of $r$ and
$p^{s = i}$, $\Pr(p^{s = i})$ denotes the marginal probability of
$p^{s = i}$. In addition, $N(r, p^{s=i})$ denotes the number of cases where $r$ and $p^{s=i}$ link the same entity pair in the KG, $N(p^{s=i})$ denotes the number of the path $p^{s=i}$ in the KG and $N(p)$ denotes the total number of paths in the KG. Since $N(p)$ can be removed from both the numerator and denominator of the fractional expression, we finally convert the probability into frequency for computation.

We filter the paths by choosing all \(p^{s = i}\) from $h$ to $t$ whose
\(\Pr\left( r \middle| p^{s = i} \right)
> 0 \). Thus, \(P^{s = i}_{r}\) is the set of all filtered \(p^{s =
i}\). Sometimes we could infer the fact not from the direct relation
${r}$ but from the path, which means the value of \(E\left( h,P^{s =
i}_{r},t \right)\) could possibly be less than that of \(E\left(
h,r,t \right)\).

Furthermore, we utilize a minimum pooling method to fuse information
from paths with different lengths and define an energy function as
follows,
\begin{equation}
\begin{split}
E_{final} & \left( h,r,t \right) = \textrm{Min} \lbrack\ E\left( h,r,t \right),\ E\left( h,P^{s = 1}_{r},t \right), \\
& E\left( h,P^{s = 2}_{r},t \right),\ldots,\ E\left( h,P^{s = n}_{r},t
\right)\rbrack, 
\label{efinal}
\end{split}
\end{equation}
where \(E\left( h,r,t \right)\) indicates the energy value of direct
relation ${r}$ and it is calculated by substituting $r$ as $p^{s=1}$
into Equation (\ref{multi-step}). \(E\left( h,P^{s = i}_{r},t \right)\)
is initialized as \textbf{infinite}, thus it will not influence the
outcome of final energy function if there is no \emph{i}-step path
between ${h}$ and ${t}$.

In summary, we adopt the min-pooling strategy twice in our model.
For\(\ E\left( h,P^{s = i}_{r},t \right)\), min-pooling aims to choose
the most matched path with ${r}$ among all \emph{i}-step paths. And
for the final energy function, min-pooling tries to extract
nonlinear features from paths of various lengths. In addition, the
min-pooling method addresses the problem that there may be no
relation paths between ${h}$ and ${t}$.

\subsection{Objective Function}

The objective function for the proposed model OPTransE is formalized
as
\begin{equation}\small
\begin{split}
 & L\left( S \right) =
 \sum_{\left( h,r,t \right) \in S}^{}\Biggl\{ L\left( h,r,t \right)\  + \lambda \cdot \sum_{i}^{}
 \biggl[\frac{1}{Z_{i}} \\
& \cdot\sum_{p^{s=i} \in P^{s=i}_{r}}^{}\Pr\left( p^{s=i} \middle| h,t
\right) \cdot \Pr\left( r \middle| p^{s=i} \right)
 \cdot L\left( h,p^{s=i},t \right)\biggr]\Biggr\},
\end{split}
\label{obj}
\end{equation}
where \(L\left( h,r,t \right)\) indicates the loss function for the
triple $(h,r,t)$, and \(L\left( h,p^{s=i},t \right)\) represents the
loss value with respect to the relation path \(p^{s=i}\). The
probability \(\Pr\left( p^{s=i} \middle| h,t \right)\) indicates the
reliability of the relation path \(p^{s=i}\) given the entity pair
$(h,t)$, and \(\Pr(r|p^{s=i})\) denotes the reliability of a path
\(p^{s=i}\) associated with the given relation ${r}$. The details of \(\Pr\left( p^{s=i} \middle| h,t \right)\)  are shown in \cite{lin2015modeling}, which is computed by a path-constraint resource allocation algorithm. \(Z_{i} =
\sum_{p^{s=i} \in P^{s=i}_{r}}^{}{\Pr\left( p^{s=i} \middle| h,t
\right)\Pr\left( r \middle| p^{s=i} \right)}\) is a normalization
factor, and \(\lambda\) is utilized to balance the triple loss and
paths losses.

We adopt the margin-based loss in our model, i.e.,
\begin{equation}\small
L\left( h,r,t \right) = \sum_{(h{'},r,t{'}) \in
S^{'}}^{}{\lbrack\gamma+E\left( h,r,t \right)-
E(h{'},r,t{'})\rbrack}_{+},
\end{equation}

\begin{equation}\small
\resizebox{1\linewidth}{!}{$ L\left( h,p^{s=i},t \right) =
\sum\limits_{(h{'},r,t{'}) \in S^{'}}^{}{\lbrack\gamma_{i}+ E\left(
h,p,t \right)- E(h{'},p,t{'})\rbrack}_{+}$},
\end{equation}
where $p$ is the simple form of $p^{s=i}$. \(\left\lbrack x
\right\rbrack_{+} = max(x,0)\) returns the higher one between $x$
and 0. \(\gamma_{i}\) is the margin to separating positive and
negative samples. It is noteworthy that we employ different margin
\(\gamma_{i}\) for paths with different step number because the
noise of energy function will be magnified as the number of steps
increases. The corrupted triple set $S{'}$ for $(h,r,t)$ is denoted
as follows:
\begin{equation}
S{'} = \{(h',r,t) \cup(h,r,t')\}.
\end{equation}
We replace the head entity or the tail entity in the triple randomly
and guarantee that the new triple is not an existing valid triple.

Our goal is to minimize the total loss. Valid relation paths will
obtain lower energy value after the optimization, so that paths can
sometimes replace directed relations when performing the prediction.

\subsection{Parameter Learning}

We utilize stochastic gradient descent (SGD) to optimize the objective
function in Equation (\ref{obj}) and learn parameters of the model. To ensure the
convergence of the model, we impose limitations to the norm of
vectors, i.e., \(||\mathbf{h}{||}_{2} \leq 1,\ ||\mathbf{r}{||}_{2}
\leq 1,||\mathbf{t}{||}_{2} \leq
1,||\mathbf{W}_{\mathbf{r,1}}\mathbf{h}{||}_{2} \leq
1,||\mathbf{W}_{\mathbf{r,2}}\mathbf{t}{||}_{2} \leq 1.\) Moreover,
we note that the objective function defined in Equation (\ref{obj})
has two parts. The first part is for the basic triple and the second
part is for the relation paths. To focus on the representation of
ordered relation paths in the second part, we only update the
parameters of relation vectors in the path when conducting the
optimization of the model.

In addition, we follow PTransE \cite{lin2015modeling} to generate reverse
relation ${r\textsuperscript{-1}}$ to enlarge the training set, and
the inference in KGs can be through the reverse paths. For instance,
for the fact (\textbf{Honolulu}, \textsl{CapitalOf},
\textbf{Hawaii}), we will also add a fact with the reverse relation
to the KG, i.e., (\textbf{Hawaii}, \textsl{CapitalOf$^{-1}$},
\textbf{Honolulu}).

\subsection{Complexity Analysis}

Let \emph{d} denote the dimension of entities and relations,
\emph{N\textsubscript{e}} and \emph{N\textsubscript{r}} denote the
number of entities and relations, respectively. The number of model
parameters for OPTransE is (\emph{N\textsubscript{e}d} +
\emph{N\textsubscript{r}d} +
2\emph{N\textsubscript{r}d\textsuperscript{2}}), which is the same
as that of STransE.

Moreover, let \emph{N\textsubscript{p}} denote the expected number
of relation paths between the entity pair, \emph{N\textsubscript{t
}} denote the number of triples for training, \emph{k} denote the
maximum length of relation paths. According to the objective
function shown in Equation (\ref{obj}) and details of parameter
learning stated in Section 3.4, the time complexity of OPTransE for
optimization is
\emph{O(k\textsuperscript{2}d\textsuperscript{3}N\textsubscript{p}N\textsubscript{t})},
which is on the same magnitude as that of RPE(MCOM)
\cite{lin2018relation}.

\section{Experiments}

\subsection{Datasets}
To evaluate the proposed model OPTransE, we use two benchmark
datasets: WN18 and FB15K as experimental data. They are subsets of
the knowledge graph WordNet \cite{miller1995wordnet} and Freebase
\cite{bollacker2008freebase}, respectively
\cite{bordes2013translating}. These two datasets have
been widely employed by researchers for KG completion
\cite{jia2018path,lin2018relation}. The statistic details of the two
datasets are shown in Table 1. In our experiments, as we add triples
of reverse relations to the datasets, the number of relations and
training triples are doubled.

\begin{table}[H]\small
\centering \caption{Statistics of datasets}
\begin{tabular}{c|ccccc}
\hline
Dataset & \#Rel & \#Ent & \#Train & \#Valid & \#Test \\
\hline
WN18 & 18 & 40,943 & 141,442 & 5,000 & 5,000 \\
FB15K & 1345 & 14,951 & 483,142 & 50,000 & 59,071 \\
\hline
\end{tabular}
\end{table}


\begin{table*}[t]
\centering \caption{Evaluation results on link prediction}
\begin{tabular}{|c|c|c|c|c|c|c|c|c|}
\hline
\multicolumn{1}{|c|}{\multirow{3}{*}{Model}} & \multicolumn{4}{|c|}{WN18}                                        & \multicolumn{4}{|c|}{FB15K}                                       \\
\cline{2-9}
\multicolumn{1}{|c|}{}                       & \multicolumn{2}{|c|}{Mean Rank} & \multicolumn{2}{|c|}{Hits@10(\%)} & \multicolumn{2}{|c|}{Mean Rank} & \multicolumn{2}{|c|}{Hits@10(\%)} \\
\cline{2-9}
\multicolumn{1}{|c|}{}                       & Raw         & Filtered        & Raw          & Filtered         & Raw         & Filtered        & Raw          & Filtered         \\
\hline
SE                                         & 1011        & 985             & 68.5         & 80.5             & 273         & 162             & 28.8         & 39.8             \\
SME                                        & 545         & 533             & 65.1         & 74.1             & 274         & 154             & 30.7         & 40.8             \\
TransE                                     & 263         & 251             & 75.4         & 89.2             & 243         & 125             & 34.9         & 47.1             \\
TransH                                     & 318         & 303             & 75.4         & 86.7             & 212         & 87              & 45.7         & 64.4             \\
TransR                                     & 238         & 225             & 79.8         & 92.0               & 198         & 77              & 48.2         & 68.7             \\
TranSparse                                 & 223         & 211             & 80.1         & 93.2             & 187         & 82              & 53.5         & 79.5             \\
STransE                                    & \underline{217}         & 206             & 80.9         & 93.4             & 219         & 69              & 51.6         & 79.7             \\
ITransF                                    & -           & \underline{205}             & -            & 94.2             & -           & 65              & -            & 81.0               \\
HolE                                        & -         & -             & -         & 94.9             & -         & -             & -         & 73.9             \\
ComplEx                                        & -         & -             & -         & 94.7             & -         & -             & -         & 84.0  \\
ANALOGY                                        & -         & -             & -         & 94.7             & -         & -             & -         & 85.4             \\
ProjE                                        & 277         & 260             & 79.4         & 94.9            & \bf{124}         & \underline{34}             & \underline{54.7}         & \underline{88.4}             \\
RTransE                                    & -           & -             & -            & -             & -           & 50              & -            & 76.2             \\
PTransE (ADD, 2-step)                                    & 235         & 221             & \underline{81.3}         & 92.7            & 200         & 54              & 51.8         & 83.4             \\
PTransE (MUL, 2-step)                                    & 243         & 230             & 79.5         & 90.9            & 216         & 67              & 47.4         & 77.7             \\
PTransE (ADD, 3-step)                                    & 238         & 219             & 81.1         & 94.2            & 207         & 58              & 51.4         & 84.6             \\
PaSKoGE                                    & -           & -             & \underline{81.3}            & 95.0             & -           & -              & 53.1            & 88.0             \\
RPE (ACOM)                                       & -         & -             & -         & -             & 171         & 41             & 52.0         & 85.5             \\
RPE (MCOM)                                       & -         & -             & -         & -             & 183         & 43             & 52.2         & 81.7             \\
RotatE      & -         & 309             & -         & \bf95.9  
& -         & 40             & -         & \underline{88.4}                                \\
\hline
OPTransE                                   & \bf211         & \bf199             & \bf83.2         & \underline{95.7}             & \underline{136}         & \bf33              & \bf58.0         & \bf89.9            \\
\hline
\end{tabular}
\end{table*}

\linespread{0.975}
\begin{table*}[]
\centering \caption{Filtered evaluation results on FB15K by mapping
properties of relations(\%)}
\begin{tabular}{|c|cccc|cccc|}
\hline
Tasks             & \multicolumn{4}{|c|}{Predicting Head Entities (Hits@10)} & \multicolumn{4}{|c|}{Predicting Tail Entities (Hits@10)} \\
\hline
Relation Category & 1-to-1       & 1-to-N      & N-to-1      & N-to-N      & 1-to-1       & 1-to-N      & N-to-1      & N-to-N      \\
\hline
SE                & 35.6         & 62.6        & 17.2        & 37.5        & 34.9         & 14.6        & 68.3        & 41.3        \\
SME (linear)      & 35.1         & 53.7        & 19.0          & 40.3        & 32.7         & 14.9        & 61.6        & 43.3        \\
SME (bilinear)    & 30.9         & 69.6        & 19.9        & 38.6        & 28.2         & 13.1        & 76.0          & 41.8        \\
TransE            & 74.6         & 86.6        & 43.7        & 70.6        & 71.5         & 49.0          & 85.0          & 72.9        \\
TransH            & 66.8         & 87.6        & 28.7        & 64.5        & 65.5         & 39.8        & 83.3        & 67.2        \\
TransR            & 78.8         & 89.2        & 34.1        & 69.2        & 79.2         & 37.4        & 90.4        & 72.1        \\
TranSparse            & 86.8         & 95.5        & 44.3        & 80.9        & 86.6         & 56.6        & 94.4        & 83.3        \\
STransE            & 82.8         & 94.2        & 50.4        & 80.1        & 82.4         & 56.9        & 93.4        & 83.1        \\
PTransE(ADD, 2-step)        & 91.0           & 92.8        & 60.9        & 83.8        & 91.2         & 74.0          & 88.9        & 86.4        \\
PTransE(MUL, 2-step)        & 89.0           & 86.8        & 57.6        & 79.8        & 87.8         & 71.4          & 72.2        & 80.4        \\
PTransE(ADD, 3-step)       & 90.1           & 92.0        & 58.7        & 86.1        & 90.7         & 70.7          & 87.5        & 88.7        \\
PaSKoGE           & 89.7           & 94.8        & 62.3        & 86.7        & 89.3         & 72.9          & 93.4        & 88.9        \\
RPE (ACOM)         & \underline{92.5}         & 96.6        & \underline{63.7}        & 87.9        & \underline{92.5}         & \underline{79.1}        & 95.1        & 90.8       \\
RPE (MCOM)           & 91.2         & 95.8             & 55.4         & 87.2             & 91.2         & 66.3             & 94.2         & 89.9             \\
RotatE          & 92.2         & \underline{96.7}             & 60.2         & \underline{89.3}             & 92.3         & 71.3             & \underline{96.1}         & \underline{92.2}             \\
\hline
OPTransE          & \bf93.1         & \bf97.4        & \bf69.0        & \bf89.8        & \bf92.8         & \bf87.4        & \bf96.7        & \bf{92.3}       \\
\hline
\end{tabular}
\end{table*}


\subsection{Experimental Settings}

We adopt the idea from TransR \cite{lin2015learning} and initialize
the vectors and matrices of OPTransE by an existing method STransE
\cite{nguyen2016stranse}. Following TransH \cite{wang2014knowledge},
\emph{Bernoulli} method is applied for generating head or tail
entities when sampling corrupted triples.

As the length of paths increases, the reliability of the path will
decline accordingly. To better determine the maximum length of paths
for experiment, before the test on FB15K, we had evaluated OPTransE
with 3-step paths on WN18. However, OPTransE (3-step) performs
comparably as OPTransE (2-step) with a higher computational cost.
This indicates that longer paths hardly contain more useful information
and it is unnecessary to enumerate longer paths. Therefore,
considering the computational efficiency, we limit the maximum
length of relation paths as 2 steps.

In our experiments, we utilize the grid search to choose the best
parameters for the two datasets, respectively. The best configurations for OPTransE are as follows: the dimension
of entity and relation vectors \(d = 50 \), the
learning rate \(\alpha = 0.0001 \), the margin \(\gamma = 5.0 \), \(\gamma_{1} = 5.0, \gamma_{2} = 5.5 \), the balance factor \(\lambda = 0.01 \) on WN18; and \(d = 100 \), \(\alpha = 0.0005 \), \(\gamma = 4.0 \), \(\gamma_{1} = 4.5, \gamma_{2} = 5.0 \), \(\lambda = 0.01 \) on FB15K. In addition, L1 norm is
employed for scoring and we run SGD for 2000 epochs in the training
procedure.

\subsection{Evaluation Metrics and Baselines}

The same as in previous work
\cite{bordes2013translating,nguyen2016stranse}, we evaluate the
proposed model OPTransE on the link prediction task. This task aims
to predict the missing entity in a triple $(h,r,t)$, i.e.,
predicting ${h}$ when ${r}$ and ${t}$ are given, or predicting ${t}$
given ${h}$ and ${r}$. When testing a fact $(h,r,t)$, we replace
either head or tail entity with all entities in the dataset and
calculate scores of generated triples according to Equation
(\ref{efinal}). And then we sort the entities with their scores in
ascending order to locate the rank of the target entity.

For specific evaluation metrics, we employ the widely used mean rank
(MR) and Hits@10 in the experiments. Mean rank indicates the average
rank of correct entities and Hits@10 means the proportion of correct
entities ranked in top 10. Higher Hits@10 or lower value of mean
rank implies the better performance of the model on the link
prediction task. Moreover, it is noted that the generated triple for
test may exist in the dataset as a fact, thus such triples will
affect the final rank of the target entity to some extent. Hence, we
could filter out these generated triples which are facts in the
dataset before ranking. If we have performed filtering, the result
will be denoted as "Filtered", otherwise it will be denoted as
"Raw".

Moreover, Bordes et al. \shortcite{bordes2013translating} defined
four categories of relations in KGs by mapping their properties such
as 1-to-1, 1-to-N, N-to-1 and N-to-N. Thus, experimental results of
distinguishing  the four different relation types have also been
recorded for comparison.

In the link prediction task, several competitive KG completion
methods are utilized as baselines, including SE
\cite{bordes2011learning}, SME \cite{bordes2014semantic}, TransE
\cite{bordes2013translating}, TransH \cite{wang2014knowledge},
TransR \cite{lin2015learning}, TranSparse \cite{ji2016knowledge},
STransE \cite{nguyen2016stranse}, ITransF
\cite{xie2017interpretable}, HolE \cite{nickel2016holographic},
ComplEx \cite{trouillon2016complex}, ANALOGY
\cite{liu2017analogical}, ProjE \cite{shi2017proje}, RTransE
\cite{garcia2015composing}, PTransE \cite{lin2015modeling}, PaSKoGE
\cite{jia2018path}, RPE \cite{lin2018relation} and RotatE \cite{sun2019rotate}. Among them,
RTransE, PTransE, PaSKoGE and RPE exploit the information of
paths between entity pairs.

\subsection{Results}
Table 2 shows the performances of different methods on the link
prediction task according to various metrics. Numbers in \textbf{bold} mean the best results among all methods and the underlined ones mean the second best. The evaluation results
of baselines are from their original work, and "-" in the table
means there is no reported results in prior work. Note that we implement ProjE and PTransE on WN18 using the public codes.

From Table 2 we
could observe that: (1) PTransE performs better than its basic model
TransE, and RPE outperforms its original method TransR. This
indicates that additional information from relation paths between
entity pairs is helpful for link prediction. Note that OPTransE
outperforms baselines which do not take relation paths into
consideration in most cases. These results demonstrate the
effectiveness of OPTransE to take advantage of the path features in
the KG. (2) OPTransE performs better than previous path-based models
like RTransE, PTransE, PaSKoGE and RPE on all metrics. This implies
that the order of relations in paths is of great importance for
reasoning, and learning representations of ordered relation paths
can significantly improve the accuracy of link prediction. Moreover,
the proposed pooling strategy which aims to extract nonlinear
features from different relation paths also contributes to the
improvements of performance.

Specific evaluation results on FB15K by mapping properties of
relations (1-to-1, 1-to-N, N-to-1, and N-to-N) are shown in Table 3.
Several methods which have reported these results are listed as
baselines. OPTransE achieves the highest scores in all sub-tasks. We
note that it is more difficult to predict head entities of N-to-1
relations and tail entities of 1-to-N relations, since the
prediction accuracy on these two sub-tasks is generally lower than
those of other sub-tasks. Surprisingly, OPTransE has achieved
significant improvements on these two sub-tasks. Especially when
predicting tail entities of 1-to-N relations, OPTransE promotes
Hits@10 to 87.4\% which is 8.3\% higher than the best performance
among baselines. Meanwhile, since the average prediction accuracy
for N-to-N relations of OPTransE on the two datasets has reached
91.1\%, we can also infer that our model has strong ability to deal
with N-to-N relations. OPTransE projects the head and tail entities
of a triple into different relation-specific spaces, thus, it is
able to better discriminate the relevant entities. Furthermore,
these results also confirm that ordered relation paths between
entity pairs which are exploited by OPTransE contain useful
information and can help to perform more accurate inference when
facing complex relations.

\section{Conclusion and Future Work}

In this paper, we propose a novel KG completion model named
OPTransE, which aims to address the issue of relation orders in
paths. In our model, we project the head entity and the tail entity
of each relation into different spaces to guarantee the order of the
path. In addition, a pooling method is applied to extract complex
and nonlinear features from numerous relation paths. Finally, we
evaluate our proposed model on two benchmark datasets and
experimental results demonstrate the effectiveness of OPTransE.

In the future, we will explore the following research directions:
(1) we will study the applications of the proposed models in various
domains, like personalized recommendation \cite{liu2018cplr}; (2) we will explore other
techniques to fuse the ordered relation information from different
paths \cite{liu2019aem}.

\section*{Acknowledgments}

This work was partially sponsored by National Key R{\&}D Program of China (grant no. 2017YFB1002000).

%
\bibliographystyle{acl_natbib}
\bibliography{scholar}

\end{document}